\documentclass{article}
\usepackage{iclr2025_conference,times}

\usepackage{amsmath,amsfonts,bm}

\def\eqref#1{equation~\ref{#1}}

\def\1{\bm{1}}

\DeclareMathAlphabet{\mathsfit}{\encodingdefault}{\sfdefault}{m}{sl}
\SetMathAlphabet{\mathsfit}{bold}{\encodingdefault}{\sfdefault}{bx}{n}

\DeclareMathOperator*{\argmax}{arg\,max}

\usepackage{hyperref}       
\usepackage{multirow}
\usepackage[utf8]{inputenc} 
\usepackage[T1]{fontenc}    
\usepackage{url}            
\usepackage{booktabs}       
\usepackage{amsfonts}       
\usepackage{nicefrac}       
\usepackage{microtype}      
\usepackage{xcolor}         
\usepackage{color}
\usepackage{tcolorbox}
\usepackage{graphicx}
\usepackage{wrapfig}
\usepackage{makecell}
\usepackage{authblk}
\usepackage{afterpage}
\usepackage{colortbl}
\usepackage{array}
\usepackage{rotating}
\usepackage{amssymb}
\usepackage{pifont}
\usepackage{xspace}
\usepackage{subcaption}
\usepackage{amsmath}
\usepackage{mathtools}
\usepackage{amsthm}

\usepackage[capitalize]{cleveref}
\crefname{section}{Sec.}{Secs.}
\Crefname{section}{Section}{Sections}
\Crefname{table}{Table}{Tables}
\crefname{table}{Tab.}{Tabs.}

\newcommand{\usc}{\raisebox{-1pt}{\includegraphics[height=0.8em]{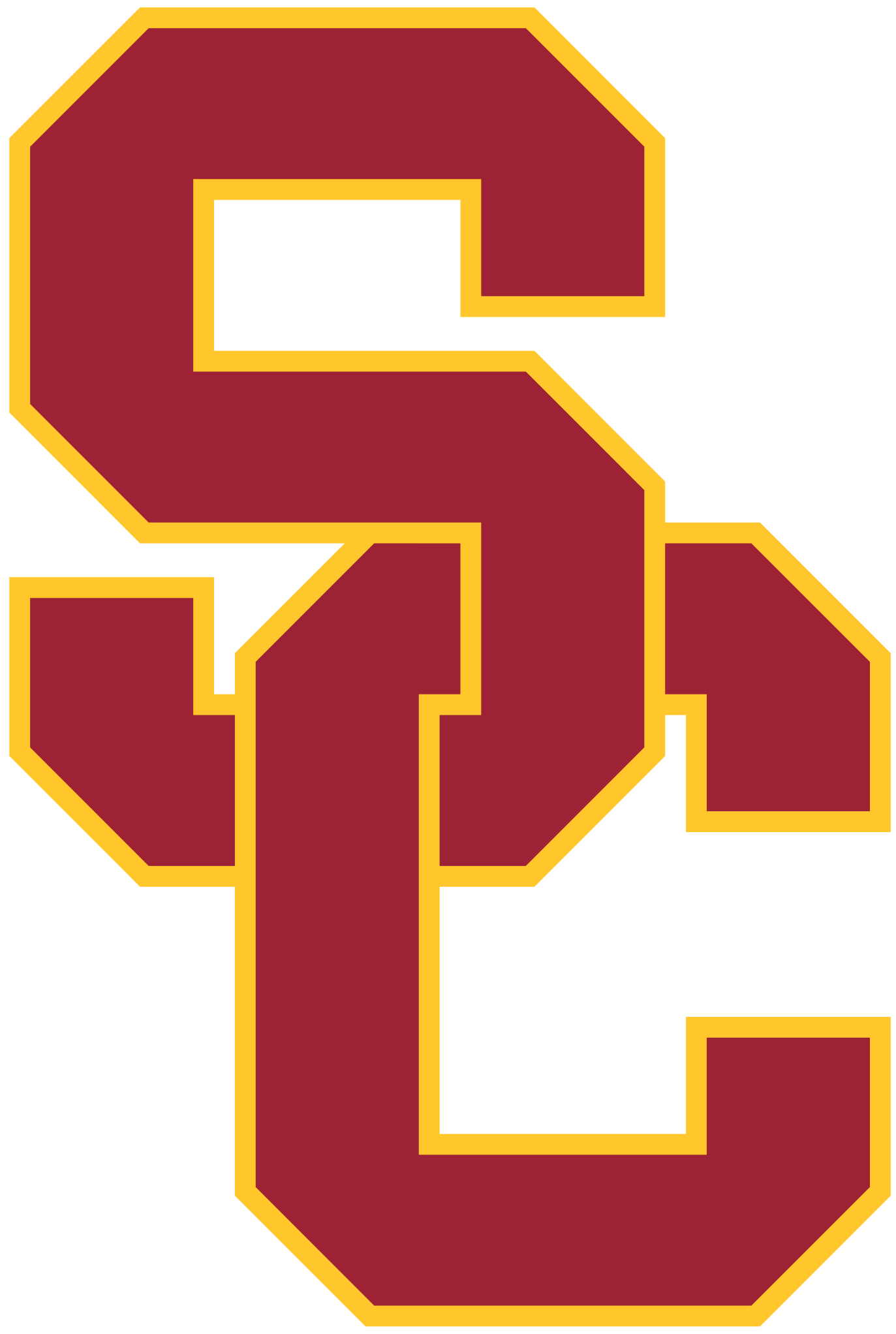}}}
\newcommand{\vuam}{\raisebox{-1pt}{\includegraphics[height=0.8em]{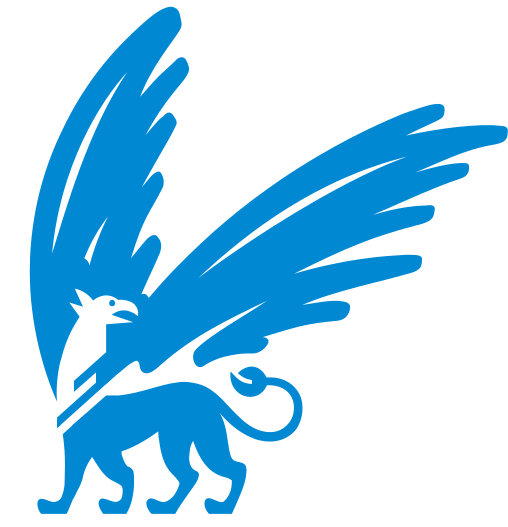}}}

\newcommand{\hc}{\texttt{human-CROP}}

\newcommand{\rel}{\texttt{rel-att}}
\newcommand{\gra}{\texttt{grad-att}}
\newcommand{\pgra}{\texttt{pure-grad}}

\makeatletter
\DeclareRobustCommand\onedot{\futurelet\@let@token\@onedot}
\def\@onedot{\ifx\@let@token.\else.\null\fi\xspace}

\def\eg{\emph{e.g}\onedot} 
\def\ie{\emph{i.e}\onedot}

\makeatletter

\definecolor{myred}{HTML}{FF8577}
\definecolor{mygreen}{HTML}{0FA958}
\definecolor{myblue}{HTML}{1982C4}
\definecolor{codegreen}{rgb}{0,0.5,0}
\definecolor{codegray}{rgb}{0.5,0.5,0.5}
\definecolor{codepurple}{rgb}{0.07,0,0.53}
\definecolor{codered}{RGB}{189,41,0}
\definecolor{codecomment}{RGB}{153,153,153}
\definecolor{backcolour}{rgb}{0.96,0.96,0.96}
\definecolor{royalblue}{rgb}{0.0, 0.14, 0.4}
\definecolor{egyptianblue}{rgb}{0.06, 0.2, 0.65}
\definecolor{royalazure}{rgb}{0.0, 0.22, 0.66}
\definecolor{portlandorange}{rgb}{1.0, 0.35, 0.21}
\definecolor{sienna}{RGB}{183,105,68}
\definecolor{saddlebrown}{RGB}{139,69,19}
\definecolor{mediumbrown}{RGB}{83,41,11}
\definecolor{darkbrown}{RGB}{58,28,7}
\hypersetup{
    colorlinks=true,
    linkcolor=sienna,
    urlcolor=royalblue,
    citecolor=royalblue,
}

\title{MLLMs Know Where to Look:\\ Training-free Perception of\\ Small Visual Details with Multimodal LLMs}

\author{Jiarui Zhang$~\usc{}$, Mahyar Khayatkhoei$~\usc{}$, Prateek Chhikara$~\usc{}$, Filip Ilievski$~\vuam{}$  \\
$\usc{}$ University of Southern California, USA\\
$\vuam{}$ Vrije Universiteit Amsterdam, The Netherlands\\}

\iclrfinalcopy 
\begin{document}

\maketitle

\begin{abstract}
Multimodal Large Language Models (MLLMs) have experienced rapid progress in visual recognition tasks in recent years. Given their potential integration into many critical applications, it is important to understand the limitations of their visual perception. In this work, we study whether MLLMs can perceive small visual details as effectively as large ones when answering questions about images. We observe that their performance is very sensitive to the size of the visual subject of the question, and further show that this effect is in fact causal by conducting an intervention study. Next, we study the attention patterns of MLLMs when answering visual questions, and intriguingly find that they consistently know where to look, even when they provide the wrong answer. Based on these findings, we then propose training-free visual intervention methods that leverage the internal knowledge of any MLLM itself, in the form of attention and gradient maps, to enhance its perception of small visual details. We evaluate our proposed methods on two widely-used MLLMs and seven visual question answering benchmarks and show that they can significantly improve MLLMs' accuracy \emph{without requiring any training}. Our results elucidate the risk of applying MLLMs to visual recognition tasks concerning small details and indicate that visual intervention using the model's internal state is a promising direction to mitigate this risk.\footnote{Our code is available at~\url{https://github.com/saccharomycetes/mllms_know}.}
\end{abstract}

\section{Introduction}
\label{sec:intro}
Multimodal large language models (MLLMs)~\citep{gpt4o,gemini,claude,qwen2vl,llavaov,kimi,r1v} have greatly progressed the state of multimodal reasoning and planning, and are rapidly being integrated into various downstream applications, ranging from robotics~\citep{llara,spatialvlm}, biomedicine~\citep{llavamed}, autonomous driving~\citep{drivegpt4,zhang2023study} to visual mathematical reasoning~\citep{gllava,mavis,euclid} and even food recipe generation~\citep{fire}. Given the rapidly growing application of MLLMs, especially in critical domains such as biomedicine and security, it is crucial to study the limitations of their visual perception to elucidate the potential risks that may affect their downstream applications.

\begin{figure*}[t]
    \centering
    \includegraphics[trim=0 0 0 0, clip, width=\textwidth]{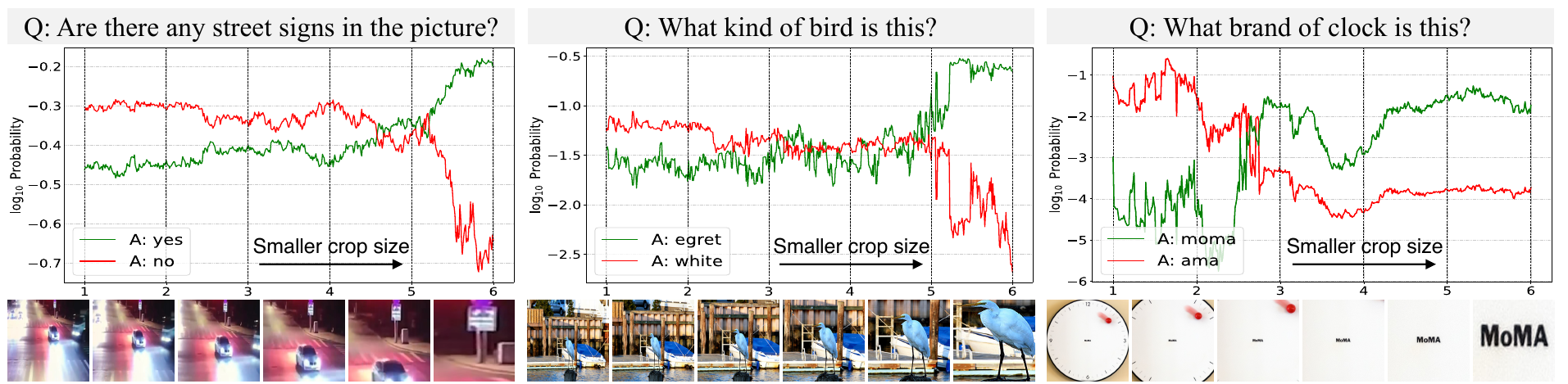}
    \caption{The effect of visual cropping on the probability of answers predicted by BLIP-2 FlanT5$_\mathrm{XL}$ zero-shot VQA model. The x-axis labels are indices to the respective cropped images displayed under each plot that the model sees at each step. The model gradually finds the correct answer.}
    \label{fig:crop_fig}
\end{figure*}

To motivate the limitation that will be the focus of this work, we start by presenting three revealing visual question answering examples in~\cref{fig:crop_fig}, in which we ask a popular MLLM BLIP-2 (FlanT5$_\mathrm{XL}$)~\citep{li2023blip} to identify an object's presence or type in each image as we vary the size of the object.
In the absence of any prior evidence, we might reasonably expect the MLLM's answer to be invariant to the size of the object, because of the MLLM's large representational capacity and pretraining on a wide variety of images containing objects of various sizes.
To the contrary, in~\cref{fig:crop_fig} (left), we observe that initially the model does not recognize the existence of a small street sign and assigns a lower probability to the correct answer; however, zooming into the image (via more focused visual cropping) towards the street sign gradually increases the probability assigned to the correct answer, suggesting that the model gradually perceives more and more relevant details of the street sign. In~\cref{fig:crop_fig} (middle), we observe further evidence of this difficulty in perceiving small details: the model initially predicts \emph{white} as the type of the bird, but when we zoom into the image towards the bird, without changing the question in any way, we observe that the model gradually assigns higher probability to the correct bird type of \emph{egret}. This suggests that the model was not making a semantic error of misunderstanding what \emph{type} means, rather it was unable to perceive sufficient details to discriminate \emph{egret} from other \emph{white} birds, which is mitigated by visual cropping. Similarly, in~\cref{fig:crop_fig} (right), we observe that the model’s initial answer is not entirely irrelevant (``ama'' compared to the correct answer ``moma''), suggesting that the model knows where to look based on the question but cannot accurately perceive the actual word, which is again mitigated by visual cropping.

In this work, we will study the limitation observed in~\cref{fig:crop_fig} extensively, elucidate its cause, and propose potential solutions to mitigate its consequences. In~\cref{sec:human_crop}, we quantitatively show that there indeed exists a difficulty in perceiving small visual concepts across various widely-used MLLMs. Our findings are consistent with prior works on evaluating the text-image matching in vision-language joint embedding models, which have observed a reverse correlation between visual object size in images and the text-image matching score~\citep{vlchecklist}, but we further establish a causal connection between visual concept size and MLLMs' perception ability through an intervention study.
In~\cref{sec:where_to_look}, we study whether the MLLMs' difficulty in perceiving small visual concepts stems from a difficulty in perceiving visual details, or from a difficulty in locating the concept due to its small size. We quantitatively show that MLLMs consistently know where to look, even when they fail to answer the question correctly.
In~\cref{sec:vicrop}, we propose three automatic visual cropping methods---leveraging the attention maps and gradients of the MLLM itself---as scalable and training-free solutions to the visual perception limitation. Finally, in~\cref{sec:experiments}, we apply our proposed methods to two popular MLLMs and evaluate them on seven visual question answering (VQA) benchmarks, showing their efficacy in enhancing MLLMs' accuracy, especially on detail-sensitive benchmarks.

\section{Related Works}
\label{sec:related_works}

\textbf{Multimodal Large Language Models~(MLLMs).}
MLLMs are foundation models capable of handling diverse language and vision tasks. These models fall into two categories: \emph{end-to-end pretrained} and \emph{modular pretrained}. End-to-end models process joint image-language data through architectures such as dual-encoder~\citep{clip}, fusion-encoder~\citep{li2021align-before-fuse}, encoder-decoder~\citep{cho2021unifying}, and unified transformer~\citep{wang2022image-as-foreign-lang}, using objectives like image-text matching, contrastive learning, and masked language modeling. Modular pretrained models, which dominate recent state-of-the-art approaches, avoid costly full pretraining by adapting existing components: 
BLIP2~\citep{li2023blip} and InstructBLIP~\citep{instructblip} train a Transformer-based connector between a frozen pretrained ViT~\citep{vit} image encoder and a frozen LLM, which transforms ViT output tokens into a fixed set of image tokens in the input space of the LLM; Qwen-VL~\citep{qwen-vl}, similarly uses a fixed-length token connector (a single cross-attention layer), but trains both the connector and the LLM; LLaVA~\citep{llava} and LLaVA-1.5~\citep{llava1.5} instead use a linear projection and a two-layer MLP as their connectors, respectively, and train both.
Our work will contribute to a better understanding of the perception limitations of MLLM and improve their perception scalably and without training, offering orthogonal benefits to existing approaches.

\textbf{Visual Localization Methods.}
Dedicated visual localization methods, such as YOLO~\citep{yolo}, SAM~\citep{kirillov2023segment}, and GLIP~\citep{glip}, rely heavily on dense spatial annotations for identifying salient image regions. Native approaches, such as Grad-CAM~\citep{gradcam}, localize regions by analyzing gradients from classifier decisions without spatial supervision. Prior work adapts Grad-CAM to BLIP~\citep{blip} leveraging its dedicated image-text similarity computation neural network called the Image-Text Matching network~\citep{pnpvqa,pnpvqa2}. In this work, we derived a more general way for localizing the attention of MLLMs to images, not relying on the specific BLIP architecture. Several recent works have explored ways to improve the visual localization capability of MLLMs for visual question answering, including chain-of-thought~\citep{visualcot,llavaplus}, tool-using~\citep{v-star}, and visual programming approaches~\citep{suris2023vipergpt,visualprogramming}. In contrast, we demonstrate that MLLMs can often effectively localize the visual subject of a question in their internal states, and propose training-free methods to leverage their internal states for improving their visual perception.

\textbf{Visual Perception Limitations in MLLMs.} The difficulty of answering questions about small objects in images has been observed by several prior and concurrent works~\citep{vicrop1,perlim,liu2024llavanext,v-star}, which have explored mitigating solutions based on high-resolution fine-tuning~\citep{liu2024llavanext,navit,qwen2vl}, multi-agent pipelines~\citep{v-star}, and use of visual cropping~\citep{vicrop1}. In this work, we provide more extensive evidence for this difficulty, establish its causal effect on MLLMs' performance, and show that it is rooted in a failure to observe small visual details as opposed to a failure to locate small objects. Several works have also shown that MLLMs suffer from object hallucination~\citep{li2023pope,rlhfv}. Furthermore, \citet{perlim} have shown visual blind spots in MLLMs---i.e., locations on the image where the MLLMs' perception degrades---as well as their sensitivity to visual quality, presence of visual distractors in the image, and even local object location perturbations.

\section{MLLMs' Sensitivity to the Size of Visual Concepts}

\begin{table*}[b!]
\caption{Sensitivity of the accuracy of MLLMs to the size of visual concepts in TextVQA. As the relative visual size of the answer decreases (right to left in each row), we observe a decline in the accuracy of the original models (no cropping) in answering questions, whereas visual cropping (\hc) significantly improves accuracy on smaller objects.}
\label{tab:bbox_size}
\centering
\begin{tabular}{llccc}
\toprule
\multirow{2}{*}{Model} & \multirow{2}{*}{Method} & \multicolumn{3}{c}{Answer Bbox Size ($S$)}\\
\cmidrule(lr){3-5}
& & \texttt{small} & \texttt{medium} & \texttt{large} \\
\midrule

\multirow{2}{*}{\makecell[c]{BLIP-2 (FlanT5$_\mathrm{XL}$)}} & no cropping 
& 12.13 & 19.57 & 36.32 \\ & \hc 
& 55.76 & 52.02 & 45.73 \\ \midrule

\multirow{2}{*}{\makecell[c]{InstructBLIP (Vicuna-7B)}} & no cropping 
& 21.79 & 30.58 & 45.30 \\ & \hc 
& 69.60 & 61.56 & 53.39 \\ \midrule

\multirow{2}{*}{\makecell[c]{LLaVA-1.5 (Vicuna-7B)}} & no cropping 
& 39.38 & 47.74 & 50.65 \\ & \hc 
& 69.95 & 65.36 & 56.96 \\ \midrule

\multirow{2}{*}{\makecell[c]{Qwen-VL (Qwen-7B)}} & no cropping 
& 56.42 & 65.09 & 68.60 \\ & \hc 
& 70.35 & 75.49 & 71.05 \\ \midrule

\multirow{2}{*}{GPT-4o} & no cropping 
& 65.76 & 72.81 & 69.17 \\ & \hc 
& 75.63 & 81.32 & 71.72 \\

\bottomrule
\end{tabular}
\end{table*}

\label{sec:human_crop}
In this section, our goal is to quantitatively study our qualitative observations in~\cref{fig:crop_fig} that MLLMs struggle with describing small visual details in images. To that end, we consider the TextVQA dataset, in which for each question we can find the image ground-truth bounding box that contains the correct textual answer. We partition its validation set into three groups based on the relative size of the ground-truth bounding box $S = \frac{A_{bb}}{A_{total}}$, where $A_{bb}$ denotes the area of the ground-truth bounding box, and $A_{total}$ the total area of the image: 1) $S<0.005$ (\texttt{small}) consisting of 773 question-image pairs, 2) $0.005\leq S<0.05$ (\texttt{medium}) consisting of 2411 question-image pairs, and 3) $S\geq 0.05$ (\texttt{large}) consisting of 1186 question-image pairs. We chose TextVQA for this study because it contains a significant number of questions about small visual concepts, and textual answers are harder for MLLMs to guess from other side information in the image (compared to object types and attributes).

\textbf{Sensitivity Study.} If a model's perception is not sensitive to the size of visual concepts, we expect it to have similar accuracy in all three partitions.  In~\cref{tab:bbox_size}, we observe that the accuracy of all MLLMs declines as the ground-truth bounding box becomes relatively smaller (right to left on the \emph{no~cropping} rows). BLIP-2 and InstructBLIP are not trained on TextVQA (\ie, are zero-shot models), and their accuracy declines by $24$ and $23$ absolute percentage points between the \texttt{large} and \texttt{small} partitions, respectively. LLaVA-1.5 and Qwen-VL are trained on the training set of TextVQA, yet, their accuracy also declines by $11$ and $12$ between the \texttt{large} and \texttt{small} partitions, respectively. Lastly, even the most recent commercial GPT-4o, with an unknown training set that might include all of TextVQA, is suffering from a $7$ percentage point decline in accuracy between small and medium partitions. These findings suggest that MLLMs have a bias against perceiving smaller visual concepts.

\textbf{Intervention Study.} The perceptual limitation we observed above might be merely correlated with size. To study whether this limitation is causally related to size, we conduct an intervention study where we provide the MLLMs with visually cropped images based on the ground-truth bounding boxes, denoted as \hc{}. More specifically, for each image-question pair and each MLLM, we crop the smallest square-shaped region containing the ground-truth bounding box from the image, and resize it to the input image resolution of the MLLM (the square-shaped cropping prevents extreme deformation of the cropped image when resizing). The cropped image is then provided to the MLLM in addition to the original image-question pair (see more details in~\cref{fig:methods}). We observe in~\cref{tab:bbox_size} that \hc{} significantly improves the accuracy of all MLLMs on the \texttt{small} and \texttt{medium} partitions, and to a lesser extent on the \texttt{large} partition. These findings show that the perception limitation is indeed caused by the size of the visual concepts, and that visual cropping can be a promising direction to mitigate this limitation.

\section{Do MLLMs Know Where to Look?}

\label{sec:where_to_look}
The limitation in perceiving small visual concepts can have two primary reasons: 1) they are hard to locate in the larger image, and 2) their small details are hard to perceive correctly. In~\cref{fig:crop_fig}, we observed that the MLLM's incorrect answer may contain partially correct information, hinting that it might know where to look in the image. In this section, we quantitatively study that observation to answer whether MLLMs' sensitivity to size is rooted in a perception limitation or a localization limitation. To that end, we first utilize the attention maps computed inside the Transformer layers of an MLLM to quantify its spatial attention over the image and then compare the total amount of this attention inside the ground-truth bounding box to other bounding boxes of the same size.

\textbf{MLLMs' Setup.} The considered MLLMs process a given image-question pair $(x,q)$ in four steps (depicted in~\cref{fig:methods}): 1)~the image is divided into $N\times N$ non-overlapping patches and processed by the ViT image encoder into $N\times N$ output tokens; 2)~the ViT output tokens are transformed into the input space of the backbone LLM---by either an MLP (LLaVA-1.5) or a Transformer connector (BLIP-2, InstructBLIP and Qwen-VL)---which we refer to as image tokens; 3)~the image tokens are then prepended to the question tokens and a predefined starting answer token, and fed to the LLM; 4) the LLM is sampled auto-regressively following the starting answer token (we use greedy sampling).

\textbf{Quantifying MLLMs' Spatial Attention over the Image.} We first measure how important each image token is to the MLLM's decision (\emph{answer-to-token attention}) by extracting the softmax cross-attention of the starting answer token to all image tokens in all layers of the backbone LLM, resulting in $A_{st}(x,q) \in \mathbb{R}^{L \times H \times 1 \times T}$, where $L, H$ are the number of layers and heads-per-layer in the LLM, and $T$ is the number of image tokens provided to the LLM. We then take the average over attention heads to arrive at the answer-to-token attention $\hat{A}_{st}(x,q) = \frac{1}{H}\sum_{h=1}^H A_{st}(x,q)$.
Next, we measure how important each image region is to each image token (\emph{token-to-image attention}). For the MLLMs that use a Transformer connector to resample ViT output tokens into a fixed number of image tokens (BLIP-2, InstructBLIP and Qwen-VL), we extract the softmax cross-attention of each image token to all ViT output tokens in all layers of the connector, resulting in $A_{ti} \in \mathbb{R}^{L_c \times H_c \times T \times N^2}$, where $L_c, H_c$ are the number of layers and heads-per-layer in the connector, $T$ the number of learnable query tokens (that become input image tokens to the LLM), and $N^2$ the number of image patches of the ViT image encoder. We then take the average over attention heads to arrive at the token-to-image attention $\hat{A}_{ti}(x) = \frac{1}{H_c}\sum_{h=1}^{H_c} A_{ti} (x)$. For LLaVA-1.5 which uses an MLP to transform ViT output tokens to image tokens, we set $\hat{A}_{ti}(x)$ to the identity tensor.
Finally, we compute the \emph{answer-to-image attention} by computing the tensor product of the answer-to-token and token-to-image attention, resulting in $A_{si}(x,q)\in \mathbb{R}^{L\times L_c \times 1 \times N^2}$ where $A^{mk}_{si}(x,q) = \hat{A}_{st}^m(x,q) \hat{A}_{ti}^k(x)$ (superscripts $m$ and $k$ denote layer indices on the LLM and the connector, respectively).

\begin{figure*}[t!]
    \includegraphics[width=\linewidth]{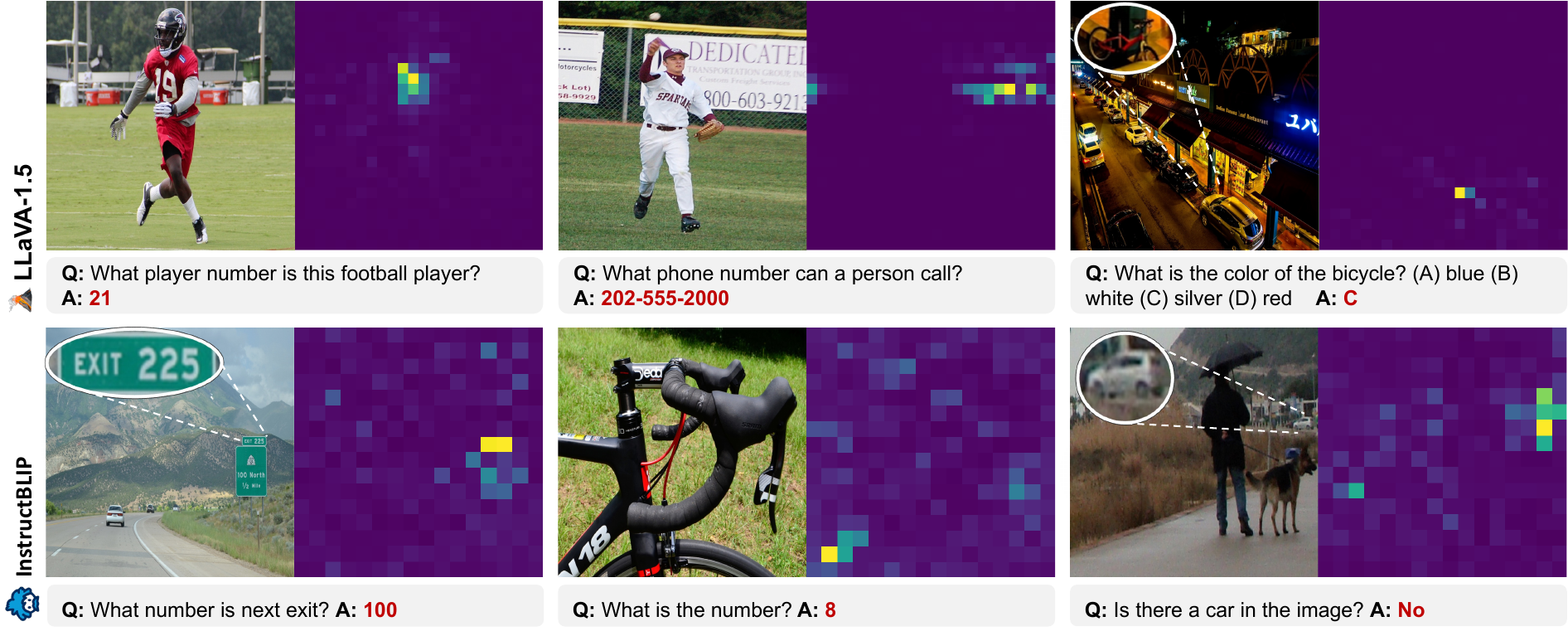}
    \caption{Examples of MLLMs knowing where to look despite answering incorrectly. The right panel in each example displays relative attention to the image (defined in~\cref{sec:where_to_look}) of one layer in the MLLM.}
    \label{fig:motivation_case}
    \vspace{-1.5em}
\end{figure*}

\begin{figure*}[b!]
    \centering
    \includegraphics[trim=0 10 0 0, clip, width=0.99\textwidth]{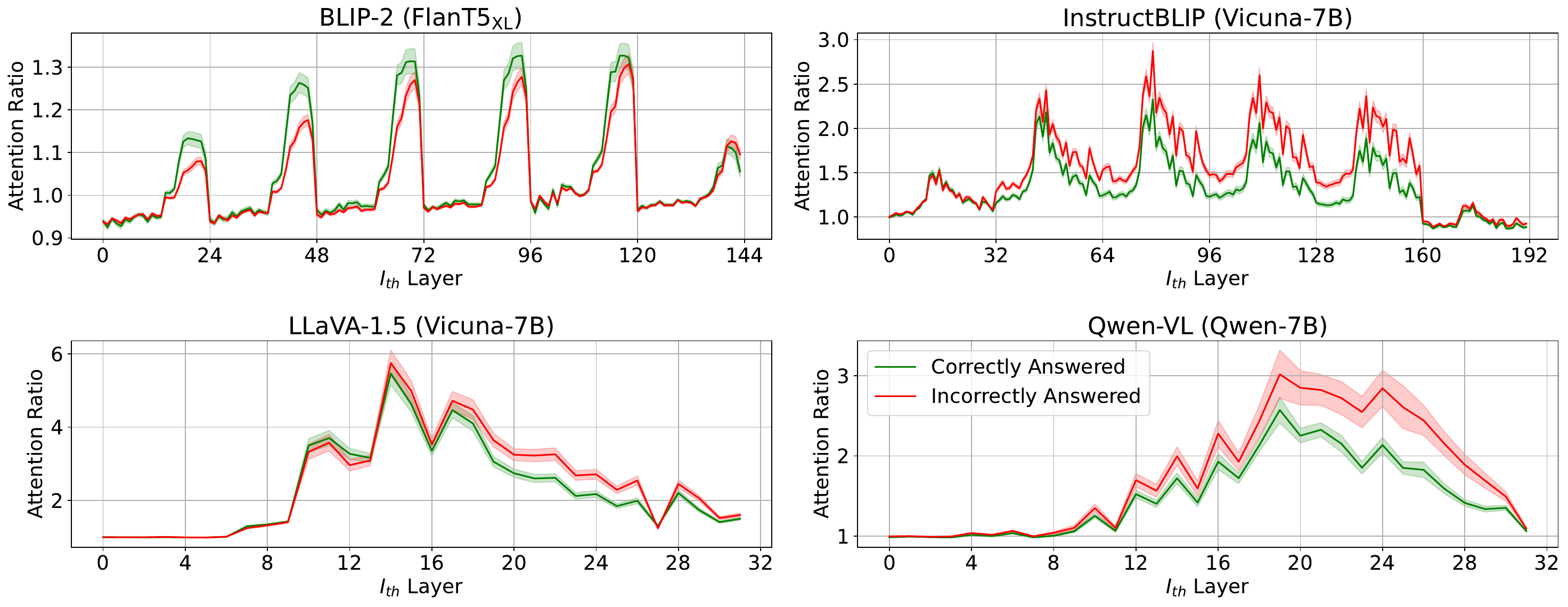}
    \caption{MLLMs' attention ratio across all layers (average with $95\%$ CI over TextVQA). The attention ratio measures how significantly the MLLM is attending to the ground-truth bounding box (defined in~\cref{sec:where_to_look}). We observe that it is greater than 1 in most layers, showing that the MLLMs know where to look in the image even when they fail to answer correctly.}
    \label{fig:where_to_look}
\end{figure*}

\textbf{Relative Attention.} One issue with using the softmax cross-attention is that not all highly attended tokens are semantically relevant to the input question. For example, recent work has observed that Transformers may use several tokens as registers to aggregate global information~\citep{registers}. To emphasize semantically relevant attention, we propose to normalize the answer-to-image attention of an image-question pair $(x,q)$ by its value on a generic instruction $q'$. Specifically, we consider a fixed instruction $q'= $``Write a general description of the image.'', and compute \textbf{relative attention} as $A_{rel}(x,q) = \frac{A_{si}(x,q)}{A_{si}(x,q')}$ under element-wise division. \cref{fig:motivation_case} shows examples of relative attention for LLaVA-1.5 and InstructBLIP ($A^{mk}_{rel}$ at layers $m=14,k=0$ and $m=15, k=2$, respectively).

\textbf{Do MLLMs Know Where to Look?}
Equipped with relative attention, we now return to our question of whether MLLMs have a localization limitation or perception limitation. To that end, we consider the validation set of TextVQA again. For each image-question pair, we first compute the relative attention. We then define \textbf{attention ratio} as the ratio of the total (sum) relative attention inside the answer ground-truth bounding box to its average across all bounding boxes of the same size as the ground-truth on the image. This ratio provides a measure of how significantly the MLLM is attending to the ground-truth bounding box (in the sense of Markov's inequality).
In~\cref{fig:where_to_look}, we plot the average (with $95\%$ confidence interval) of the attention ratio, over the validation set of TextVQA for all layers in the considered MLLMs. The horizontal axis shows the combined layer index $l = m + k \times L$ for $m \in \{0\dots L-1\}$ spanning the number of cross-attention layers in the backbone LLM, and $k \in \{0\dots L_c-1\}$ spanning the number of cross-attention layers in the connector (BLIP-2: $L=24, L_c=6$; InstructBLIP: $L=32, L_c=6$; Qwen-VL: $L=32, L_c=1$; LLaVA-1.5: $L=32, L_c=1$).
In all MLLMs, we observe a significantly larger than 1 attention ratio in most layers, suggesting that the models are attending significantly to the ground-truth bounding box region on the image. Intriguingly, the models show similarly strong attention to the correct region regardless of whether they can answer the question correctly or incorrectly. These observations show that the MLLMs tend to know where to look, even when they answer incorrectly.

\section{Automatic Visual Cropping (ViCrop)}
\label{sec:vicrop}

\begin{figure*}[b]
    \centering
    \includegraphics[trim=0 0 0 0, clip, width=0.8\textwidth]{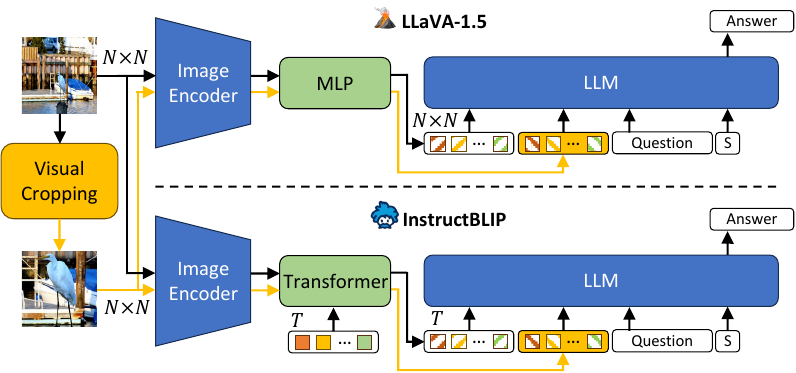}
    \caption{Illustration of the proposed visual cropping approach applied to two MLLMs.}
    \label{fig:methods}
\end{figure*}

We observed in~\cref{sec:where_to_look} that the sensitivity of MLLMs to visual concept size is primarily a perception limitation (rather than a localization limitation). Therefore, one solution to mitigate this limitation is to simply train MLLMs with a larger number of image patches while maintaining per-patch resolution (hence increasing the image resolution of MLLMs). However, increasing the input image resolution by a factor of $\alpha$, increases the number of ViT input patches (and output tokens) from $N^2$ to $\alpha^2 N^2$, which in turn increases the softmax attention computation complexity on the order of $\alpha^4 N^4$. Given that this solution is not scalable for current Transformer-based MLLMs, we choose to explore an alternative solution that \textbf{does not require any training and is scalable to any image resolution}. We note that several concurrent works have explored the first direction of \textit{training} MLLMs with higher resolution image patches~\citep{li2024mini-gemini,sun2024parrot,li2024monkey,mckinzie2024mm1,xu2024llava-uhd,luo2024feast}, and notably LLaVA-Next~\citep{liu2024llavanext} has achieved the VQA state-of-the-art in several VQA benchmarks at the time of writing. We believe our work is parallel to these efforts in the following sense: rather than training higher and higher resolution MLLMs to enable them to see all resolutions (which is inevitably upper bounded), we explore how to smartly adjust the input image towards what an MLLM already can see without any additional training. We provide evidence showing that our training-free method can provide orthogonal benefits to the training-based methods in~\cref{app:llavanext,app:seal}.

Inspired by our findings that MLLMs tend to know where to look (\cref{sec:where_to_look}) and that visual cropping can mitigate the perception limitation (\cref{sec:human_crop}), in this section we construct three automatic visual cropping methods in order to mitigate the perception limitation of MLLMs. These methods seek to use the internal information of an MLLM itself---in the form of attention maps and gradients---to find the approximate region of interest in images (\ie, the region containing the subject of a question), and then to zoom into that region via visual cropping.
One potential drawback of visual cropping is that some questions might need to have a global view of the image. To address this issue, we utilize the fact that MLLMs typically convert the image into a series of tokens. This allows us to directly extend the original image tokens by concatenating the visually cropped image tokens, as illustrated in~\cref{fig:methods}. We use this concatenation approach when applying all our methods to MLLMs.

\textbf{Relative Attention ViCrop (\rel{}).} In this method, we directly compute the relative attention $A_{rel}(x,q)$ defined in~\cref{sec:where_to_look} for each image-question pair $(x,q)$. We then select a target layer (in LLM and connector) based on a small held-out set of samples in TextVQA and use its relative attention as the importance map for visual cropping (discussed below). We ablate on the choice of layer in~\cref{sec:experiments}.

\textbf{Gradient-Weighted Attention ViCrop (\gra{}).} The relative attention runs an additional generic instruction through the MLLM to normalize the answer-to-image attention and emphasize semantically relevant attention. As an alternative that does not require a second forward pass, we consider using the gradients to normalize attention, because the gradient of the model's decision with respect to an attention score shows how sensitive the decision is to changes in that attention, hence how semantically relevant the attention is for answering the question.
To get a differentiable representation of the model's decision, we consider the logarithm of the maximum output probability at the starting answer token, $v = \log \text{softmax}(z(x,q))_{t^*} \in \mathbb{R}$, where $z \in \mathbb{R}^{D}$ is the output logit of the LLM at the starting answer position, $D$ the vocabulary size, and $t^* = \argmax_t z_t$. Then, following our notation in~\cref{sec:where_to_look}, we can compute the gradient-weighted versions of answer-to-token attention $\tilde{A}_{st}(x,q) = A_{st}(x,q) \odot \sigma(\nabla_{A_{st}}v(x,q))$ and token-to-image attention $\tilde{A}_{ti}(x,q) = A_{ti}(x) \odot \sigma(\nabla_{A_{ti}}v(x,q))$, where $\odot$ is element-wise product and $\sigma(w)=\max(0, w)$. We remove negative gradients because they correspond to tokens that if attended to will reduce the model certainty, hence often distracting locations~\cite{gradcam}. Finally, we compute the gradient-weighted answer-to-image attention as the following tensor product $\tilde{A}_{si}(x,q) = \tilde{A}_{st}(x,q) \otimes \tilde{A}_{ti}(x,q) \in \mathbb{R}^{L \times L_c \times 1 \times N^2}$. We will select the same target layer identified in~\rel{} from $\tilde{A}_{si}(x,q)$ as the importance map for visual cropping.

\textbf{Input Gradient ViCrop (\pgra{}).} In this method, we seek to find the relevant regions on the image directly using the gradient of the MLLM's decision with respect to the input image. Compared to the previous attention-based methods, \pgra{} is more versatile because it does not rely on the Transformer-based architecture. Specifically, for each image-question pair $(x,q)$, we will compute $G(x,q) = \lVert \nabla_x v(x,q) \lVert_2$, where $v(x,q)$ is the logarithm of the maximum output probability of the LLM at the starting answer token (as defined in \gra{} above), and the L2-norm is taken over the image channel dimension. However, gradients sometimes show high values in entirely constant-color regions (\eg, blue skies). Given that these non-edge regions do not contain any visual details, we will explicitly diminish them in $G$. To that end, we first apply a $3\times 3$-size Gaussian high-pass filter to the image, followed by a median filter of the same size to reduce salt-and-pepper noise. The resulting filtered image is then thresholded at its spatial median value to become a binary mask and is element-wise multiplied by $G$. Finally, the edge-emphasized $G$ is spatially average-pooled into the $N\times N$ patches of the MLLM to become an importance map for visual cropping.

\textbf{Bounding Box Selection for Visual Cropping.}
To convert the importance map (from each of the methods described above) to a bounding box, we use sliding windows of different sizes inspired by object detection literature~\cite{yolo}. Specifically, for each MLLM, we define a set of windows with sizes equal to a multiple of the input image resolution of the MLLM. The multiples are in $\{1, 1.2, \dots 2\}$. We slide each window over the image with a stride of 1 and find its best position where the sum of the importance map inside the window is maximized. After selecting the best position per window, we choose the window whose internal sum has the largest difference from the average internal sum of its adjacent positions. This latter step is a heuristic to avoid choosing too small or too large windows (notice that in both cases, moving the window slightly left/right or up/down will not change its internal sum significantly). The chosen window is then cropped from the image, resized to the input image resolution of the MLLM, and provided to the MLLM in addition to the image-question pair.

\textbf{High-Resolution Visual Cropping.} In one of the benchmarks we consider in this work, V$^*$~\cite{v-star}, the images are of very high resolution (always more than 1K) and consequently, the resized input image provided to the MLLM might completely lose the visual concept of interest for a question. To mitigate this, on this benchmark, we employ a two-stage strategy. In the first stage, we divide images into non-overlapping blocks of smaller than $1024\times 1024$ with an aspect ratio close to 1, compute the importance map separately for each block according to the ViCrop methods, and then re-attach the blocks back together. In the second stage, we find the bounding box for visual cropping on this re-attached importance map exactly as described before and provide the original image-question pair with the resized cropped image to the MLLM.

\section{ViCrop Method Analysis}
\label{sec:experiments}

In this section, we apply our proposed visual cropping methods to two open-source SOTA MLLMs, InstructBLIP (Vicuna-7B)~\citep{instructblip} and LLaVA-1.5 (Vicuna-7B)~\citep{llava1.5}. We evaluate their effectiveness in improving the perception of smaller visual concepts on 4 detail-sensitive datasets (TextVQA
\footnote{$^\dagger$In TextVQA evaluation, we do not provide externally extracted OCR tokens to the MLLM since we want to measure its true perception, this differs from the setup in the original paper. See more discussion in~\cref{app:imp}.}
~\citep{textvqa}, V$^*$~\citep{v-star}, POPE~\citep{li2023pope}, DocVQA~\citep{docvqa}), and their ability to maintain performance on larger visual concepts in 3 general-purpose datasets containing mostly large objects (GQA~\citep{hudson2019gqa}, AOKVQA~\citep{schwenk2022okvqa}, VQAv2~\citep{goyal2017vqav2}). InstructBLIP uses the hyper-parameters $N=16, m=15, k=2$ and input image resolution of $224\times 224$. LLaVA-1.5 uses $N=24, m=14$ and input image resolution of $336\times 336$.  When reporting accuracy, we compute VQA-score\footnote{https://visualqa.org/evaluation.html} for all benchmarks except GQA. For GQA, we compute accuracy using the official code.\footnote{https://cs.stanford.edu/people/dorarad/gqa/evaluate.html}. See~\cref{app:imp,app:dataset,app:zeroshot_instruction} for further details about implementation, datasets, and prompts.

\begin{figure*}[t]
    \centering
    \includegraphics[trim=0 0 0 0, clip, width=\textwidth]{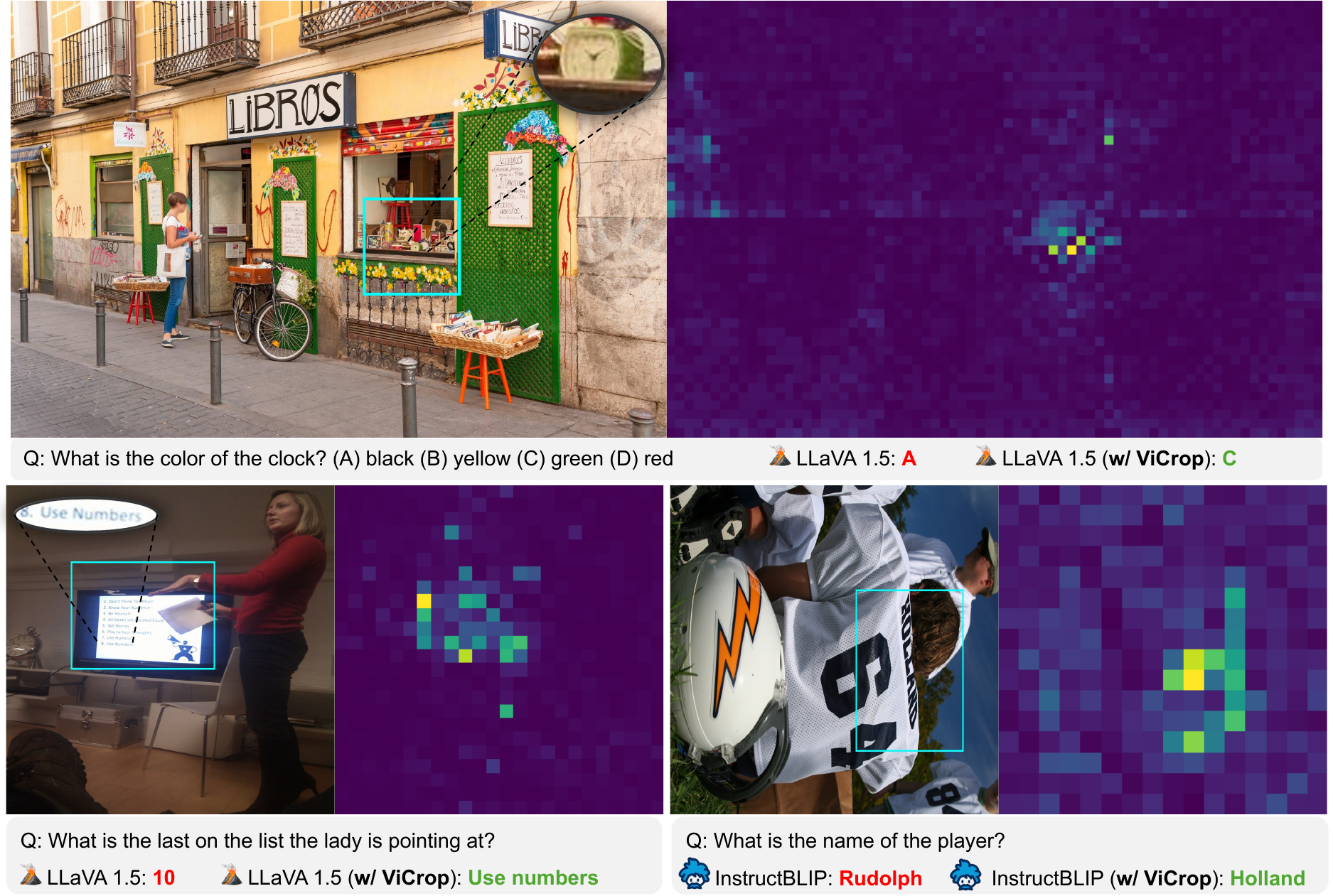}
    \caption{Examples of \rel{} helping MLLMs correct their mistakes (cyan-colored bounding box shows cropped region by \rel{}; zoom-in insets are displayed for better readability).}
    \label{fig:crop_examples}
\end{figure*}

\begin{table*}[!t]
\caption{Accuracy of the proposed ViCrop methods on visual question answering benchmarks.}
\label{tab:main_result}
\centering
\scalebox{0.9}{
\begin{tabular}{l l c c c c c c c}
\toprule
\multirow{2}{*}{Model} & & \multicolumn{4}{c}{Smaller Visual Concepts} & \multicolumn{3}{c}{Larger Visual Concepts} \\
\cmidrule(lr){3-6} \cmidrule(lr){7-9}
& & TextVQA$^\dagger$ & V* & POPE & DocVQA & AOKVQA & GQA & VQAv2 \\
\midrule
\multirow{4}{*}{\makecell[c]{LLaVA-1.5}} & no cropping 
& 47.80      & 42.41      & 85.27            & 15.97          & 59.01          & 60.48          & 75.57          \\ & \rel{}
& 55.17      & \bf{62.30} & \textbf{87.25}   & 19.63          & \textbf{60.66} & 60.97          & \textbf{76.51} \\ & \gra{}
& \bf{56.06} & 57.07      & 87.03            & \textbf{19.84} & 59.94          & \textbf{60.98} & 76.06          \\ & \pgra{}
& 51.67      & 46.07      & 86.06            & 17.70          & 59.92          & 60.54          & 75.94          \\
\midrule
\multirow{4}{*}{\makecell[c]{InstructBLIP}} & no cropping
& 33.48          & 35.60          & 84.89          & 9.20          & 60.06          & 49.41          & 76.25          \\ & \rel{} 
& 45.44          & \textbf{42.41} & 86.64          & 9.95          & 61.28          & 49.75          & \textbf{76.84} \\ & \gra{}
& \textbf{45.71} & 37.70          & \textbf{86.99} & \textbf{10.81}& \textbf{61.77} & \textbf{50.33} & 76.08          \\ & \pgra{}
& 42.23          & 37.17          & 86.84          & 8.99          & 61.60          & 50.08          & 76.71          \\
\bottomrule
\end{tabular}
}
\end{table*}

\textbf{ViCrop Improves VQA Accuracy.} In \cref{fig:crop_examples}, we show a few examples of the ViCrop helping the MLLM correct itself (more examples in~\cref{app:examples}), and in~\cref{tab:main_result}, we report the accuracy of the proposed ViCrop methods on the VQA benchmarks. We observe that all methods significantly improve the accuracy of the original MLLMs (\emph{no cropping}) on detail-sensitive benchmarks, without requiring any training, while maintaining the MLLMs' performance on benchmarks with larger visual concepts. Thus, the accuracy gain on fine details (most notably in TextVQA and V$^*$) does not seem to come at the cost of accuracy on larger visual details and relations. We also observe that the accuracy gain for LLaVA-1.5 is more substantial than for InstructBLIP. This can be explained by the fact that InstructBLIP only trains its connector and not its backbone LLM during tuning---the LLM does not adapt to use the image tokens, rather the image tokens are adapted to optimally prompt the LLM---and therefore the LLM cannot effectively use the additional (cropped) image tokens provided through visual cropping. Nonetheless, the results show that ViCrop can be effectively applied to different MLLMs, and is a promising inference-time solution for mitigating the perception limitation observed in~\cref{sec:human_crop}.

\begin{table*}[!t]
\caption{Ablation study on the choice of layer and the use of high-resolution visual cropping.}
\label{tab:layer_res}
\centering
\scalebox{0.9}{
\begin{tabular}{l l c c c c c c}
\toprule
\multirow{2}{*}{Model} & & \multicolumn{3}{c}{Choice of Layer} & \multicolumn{3}{c}{High-Resolution ViCrop} \\
\cmidrule(lr){3-5} \cmidrule(lr){6-8}
& & Selective & Average & $\Delta$ & w/ High-Res & w/o High-Res & $\Delta$ \\
\midrule
\multirow{4}{*}{\makecell[c]{LLaVA-1.5}} & no cropping 
& 47.80 & -- & -- & 42.41 & 42.41 & -- \\ & \rel
& 55.17 & 55.45 & +0.28 & 62.30 & 47.64 & -14.66 \\ & \gra
& 56.06 & 56.26 & +0.20 & 57.07 & 49.74 & -7.33 \\ & \pgra
& 51.67 & -- & -- & 46.07 & 45.03 & -1.04 \\
\midrule
\multirow{4}{*}{\makecell[c]{InstructBLIP}} & no cropping
& 33.48 & -- & -- & 35.60  & 35.60 & -- \\ & \rel
& 45.44 & 44.40 & -1.04 & 42.41 & 38.74 & -3.67 \\ & \gra
& 45.71 & 44.98 & -0.73 & 37.70 & 42.41 & +4.71 \\ & \pgra
& 42.23 & -- & -- & 37.17 & 42.41 & +5.24 \\
\bottomrule
\end{tabular}
}
\end{table*}

\textbf{Ablation Study on the Choice of Layer.}
To understand the importance of the choice of an informative layer for \rel{} and \gra{} (as discussed in~\cref{sec:vicrop}), in~\cref{tab:layer_res} we compare the accuracy of these methods when simply taking the average of all layers in ${A_{rel}}$ and $\tilde{A}_{si}$, respectively, on TextVQA. We observe that \rel{} is robust to this choice and \gra{} declines about $3.5$ percentage points in accuracy. Importantly, both methods still improve the MLLMs' accuracy even when using the layer average, suggesting that averaging is a suitable choice in the absence of any data for selecting a layer.

\textbf{Ablation Study on the High-Resolution ViCrop.} In~\cref{sec:vicrop}, we proposed a two-stage strategy for processing the very high-resolution images in the V$^*$ benchmark. To see how effective this strategy is, in~\cref{tab:layer_res} we compare the accuracy of ViCrop methods with and without this high-resolution strategy on V$^*$. We observe that while this strategy is very beneficial to LLaVA-1.5, it declines the performance of \gra{} and \pgra{} for InstructBLIP. However, all methods, with and without this strategy, still improve the MLLMs' accuracy.

\textbf{ViCrop with External Tools.}
In addition to the internal ViCrop methods, we also considered the use of external off-the-shelf models to find the region of interest in an image for visual cropping. Specifically, we utilized SAM~\citep{kirillov2023segment}, YOLO~\citep{yolo}, and CLIP~\citep{clip} to find the most relevant part of an image to a given question (details of these external ViCrop methods are provided in~\cref{app:external}). In~\cref{tab:externals_time}, we compare the accuracy of external ViCrop methods to the internal methods on TextVQA. While external models are also effective in improving the accuracy of MLLMs, they are weaker than all the proposed internal ViCrop methods, thus we did not explore them further.

\begin{table*}[!t]
\caption{Accuracy of ViCrop using external tools compared to attention/gradient (on TextVQA); and the inference time overhead of ViCrop methods (in seconds). Original's time is per answer token.}
\label{tab:externals_time}
\centering
\scalebox{0.85}{
\begin{tabular}{l l c c c c c c c}
\toprule

& Model & Original & SAM & YOLO & CLIP & \rel{} & \gra{} & \pgra{} \\
\midrule
\multirow{2}{*}{\makecell[c]{Accuracy\\(TextVQA)}} & LLaVA-1.5 & 
47.80 & 49.42 & 48.84 & 48.55 & 55.17 & 56.06 & 51.67 \\ 
& InstructBLIP & 33.48 & 39.23 & 36.49 & 39.61 & 45.44 & 45.71 & 42.23 \\
\midrule
\multirow{2}{*}{\makecell[c]{CPU Time}} & LLaVA-1.5 & 
2.26 & \multirow{2}{*}{\makecell[c]{91.53}} & \multirow{2}{*}{\makecell[c]{0.97}} & \multirow{2}{*}{\makecell[c]{5.46}} & 14.43 & 11.33 & 29.86 \\ 
& InstructBLIP & 0.66 &  &  &  & 4.35 & 3.78 & 7.04 \\
\midrule
\multirow{2}{*}{\makecell[c]{GPU Time}} & LLaVA-1.5 & 
0.17 & \multirow{2}{*}{\makecell[c]{3.33}} & \multirow{2}{*}{\makecell[c]{0.35}} & \multirow{2}{*}{\makecell[c]{1.07}} & 1.16 & 0.89 & 2.36 \\ 
& InstructBLIP & 0.06 &   &   &   & 0.28 & 0.29 & 0.60 \\
\bottomrule
\end{tabular}
}
\end{table*}

\textbf{Inference Time Overhead.}
In~\cref{tab:externals_time}, we report the average inference-time overhead of the proposed visual cropping methods on GPU (NVIDIA RTX A6000) and CPU (Intel(R) Gold 5317 CPU @ 3.00GHz) and compare with the per-answer-token processing time of the MLLMs.
We see that all proposed methods (except SAM) are reasonably fast (1 to 2 seconds overhead on GPU). For example, computing the visual cropping with \rel{} takes the time of generating only 5 tokens by the MLLM. \textbf{Note that our methods’ time overhead will not scale with the number of answer tokens and is constant regardless of how long the answer is} because our external methods do not need any answer token, and internal methods only need the starting answer token (see~\cref{sec:vicrop}). In contrast, MLLMs’ inference time scales approximately linearly with the number of answer tokens.

\section{Conclusion}
In this work, we qualitatively and quantitatively showed that there exists a perception bias against small visual details in widely-used MLLMs. Then we found that MLLMs often know where to look even if they fail to answer the question, indicating that the bias toward small visual details is rooted in a perception limitation rather than a localization limitation. To mitigate this limitation, we proposed multiple automatic visual localization methods as scalable and training-free solutions based on models' internal dynamics while answering the visual questions. Through evaluation of multiple multimodal benchmarks, we showed that our method can significantly improve MLLMs’ accuracy without requiring any training, especially in detail-sensitive scenarios. Our findings suggest that MLLMs should be used with caution in detail-sensitive applications, and that visual cropping/localization with the model's own knowledge is a promising direction to enhance their performance.

\textbf{Limitations and Future Work.}
The proposed ViCrop methods do not enhance all types of questions equally. We have observed that questions concerning relations and counting are particularly difficult for ViCrop methods to help answer. This is expected as the proposed ViCrop can only focus on one region in the image. We leave extending ViCrop to focus on multiple regions simultaneously for future work. Another limitation of the proposed methods is their time overhead and the addition of visual tokens. While the overhead is reasonable (a few seconds), we believe it can be significantly optimized as an inference-time mechanism, for example by utilizing lower precision, and weight quantization. Furthermore,
Matryoshka Query Transformer (MQT)~\citep{matryoshka} enables MLLMs to have varying visual context size during inference. In our current results, we have shown that our methods can work with two different MLLMs with distinct visual context sizes, so it seems entirely possible that our method can still work with varying visual context size under MQT, which can further reduce our computational cost through rescaling the cropped image. We leave these inference cost optimizations to future works. Lastly, we have observed that the proposed methods tend to have some complementary benefits, and therefore exploring ways to combine them (for example based on the prediction uncertainty) is also a very interesting direction for future research.

\clearpage

\subsubsection*{Acknowledgments}

We thank Jinyi Hu and Joe Mathai for their very useful insights. We also express our gratitude to anonymous reviewers for their valuable feedback. This research was supported in part by the National Science Foundation under Contract No. IIS-2153546.

\bibliography{cite}
\bibliographystyle{plainnat}

\clearpage
\appendix
\section{Implementation Details}
\label{app:imp}
We use \textit{python 3.10.6, transformers 4.29.1 and torch 2.1.2} for all the experiments. Our environment consists of an Intel(R) Gold 5317 CPU @ 3.00GHz with 48 cores and 756 GB of RAM, and we utilize NVIDIA RTX A6000 GPUs for our experiments. We use the huggingface implementations of all studied MLLMs with the recommended hyper-parameters according to the respective papers. For GPT-4o, we use the official public API, which is available at the time of submission.

Regarding the evaluation setting of the TextVQA dataset in~\cref{tab:main_result}, our setting is slightly different from the one used by the LLaVA-1.5 original paper~\cite{llava1.5}. They report accuracy on TextVQA by using externally extracted OCR tokens to enrich its text prompt. This is a text-specific trick that essentially out-sources the perception of text to an external OCR model. This text-specific trick is not mentioned in their paper or supplementary material, but see their clarification in response to a GitHub issue here: \url{https://github.com/haotian-liu/LLaVA/issues/515#issuecomment-1763779341}. In contrast, we treat TextVQA the same as any other vision dataset in our experiments, that is, we do not provide any OCR extracted tokens to MLLMs when applying them to TextVQA (only image and question, in the evaluation prompt format specified in their respective papers). This results in a slightly lower accuracy compared to the one reported in the original paper, but instead, this number shows the true perception ability of LLaVA-1.5 on TextVQA, not confounded by the ability of an external OCR model. For completeness, we also measured TextVQA accuracy in the presence of OCR tokens, which results in $59.8$ for LLaVA-1.5 without any visual cropping, and $63.95$ with \rel{}, showing that our proposed visual cropping can still be beneficial even when OCR tokens are provided to the MLLM.

\section{Dataset Statistics}
\label{app:dataset}
In this section, we present the details of the datasets used for evaluation in this paper. We report the average height and weight of the images in the dataset. We also report the number of images and questions in each dataset. For VQAv2, we run our experiment on a random 50K subset of the official validation set. We use the entire validation set in all other datasets.

\begin{table}[h]
\caption{Average width ($\bar{W}$) and height ($\bar{H}$) of images, number of images, and number of questions on all datasets.}

\centering
\begin{tabular}{lccccccc}
\toprule
 & V$^{*}$ & DocVQA & TextVQA & POPE & AOKVQA & GQA & VQAv2 \\
\midrule
$\bar{W}$  & 2246 & 1776 & 954 & 584 & 581 & 578 & 577\\ 
$\bar{H}$ & 1582 & 2084 & 818 & 478 & 480 & 482 & 485\\ 
\# Images & 191 & 1286 & 3166 & 500 & 1122 & 398 & 14206 \\
\# Questions & 191 & 5349 & 5000 & 8910 & 1145 & 10781 & 50000 \\

\bottomrule
\end{tabular}
\label{tab:example}
\end{table}

For our analysis presented in Table \ref{tab:bbox_size} and Figure \ref{fig:where_to_look}, we focused on TextVQA dataset, which includes bounding box annotations for OCR-detected text within images. However, this dataset does not specify which bounding boxes correspond to the regions where answers are located, necessitating a manual annotation process.
The TextVQA dataset comprises 5000 questions and 3166 images. We manually annotated these question-image pairs, ensuring accurate bounding boxes over \textbf{all the regions} of interest where the answers could be found.
This manual annotation process was essential for our analysis, allowing us to provide precise and reliable ground-truth data for the study.
Given that some questions were associated with multiple bounding boxes in their corresponding images, we undertook a filtering process to isolate the question-image pairs. This effort resulted in a refined set of 4370 question-image pairs, where there is only one instance of the subject of the question in the image. For example, if the question is ``what type of drink is sold here?'' and there are two different cans of drinks in the image, we remove this image-question pair.

\section{Prompt Format for Zero-shot Inference}
\label{app:zeroshot_instruction}

In this section, we provide details about the prompt format used in models for zero-shot inference. We use a different prompt format for LLaVA and InstructBLIP which we adapt from the original papers, as shown below.

\begin{tcolorbox}
\textbf{LLaVA-1.5}\\

\texttt{<image>
USER:\{question\} Answer the question using a single word or phrase.
ASSISTANT:}
\end{tcolorbox}

\begin{tcolorbox}
\textbf{InstructBLIP} \\ 

\texttt{<image> Question:\{question\} Short Answer:}
\end{tcolorbox}

\section{Orthogonal Benefits to LLaVA-NeXT}
\label{app:llavanext}
We apply our proposed \rel{} visual cropping method to an additional newer MLLM -- LLaVA-NeXT~\citep{liu2024llavanext} current SOTA in several VQA benchmarks -- that has support for higher-resolution compared to LLaVA-1.5. In~\cref{tab:llavanext}, we observe that our method can still boost the MLLM's performance, without requiring any training. This provides further evidence for the generalizability of our proposed visual cropping and its orthogonal benefits to training MLLMs with higher image patch resolution.

\begin{table}[h]
\caption{Orthogonal benefits of visual cropping when applied to LLaV-NeXT that is trained to adapt to processing high-resolution images.}

\centering
\begin{tabular}{lcc}
\toprule
Model & TextVQA & V$^{*}$ \\
\midrule
LLaVA-NeXT (Mistral-7B) & 65.17 & 58.11\\
LLaVA-NeXT (Mistral-7B) + \rel{} & 68.65 & 61.78\\

\bottomrule
\end{tabular}
\label{tab:llavanext}
\end{table}

\section{Comparison with the V* method (SEAL)}
\label{app:seal}
The V* method (SEAL)~\citep{v-star} proposes a multi-agent fine-tuning approach to enhance the ability of an underlying MLLM to answer questions about small visual concepts.
However, SEAL requires substantial training and finetuning of several neural networks, whereas our methods are completely training-free, so a direct comparison would not be fair. Nonetheless, to provide an idea of how our method compares to SEAL in an “as-is” fashion (i.e. if a user just wants to pick one method as-is off-the-shelf), we report the accuracy of SEAL compared to LLaVA-1.5+\rel{} in~\cref{tab:seal_comparison}. We observe that our method outperforms SEAL except on the V* benchmark. We think this might be because SEAL is designed and tuned specifically toward high-resolution images in its V* benchmark. We also note that the inference time of SEAL is slower than our method (4.44s compared to 1.88s on average per question, tested on the same random 100 TextVQA samples with one A6000 GPU).
That being said, we note that our methods and SEAL can both help enhance MLLMs, and our methods can be integrated into SEAL or other multi-agent pipelines.

\begin{table}[h] 
\small
\caption{Performance comparison between our~\rel{} applied on LLaVA-1.5 and SEAL~\citep{v-star} across multiple vision-language benchmarks.} 
\centering 
\begin{tabular}{lccccccc} 
\toprule Model & TextVQA & V$^*$ & POPE & DocVQA & AOKVQA & GQA & VQAV2 \\
\midrule SEAL  & 36.30 & 75.30 & 82.40 & 5.31 & 55.34 & 50.18 & 65.35 \\
LLaVA-1.5+\rel{}  & 55.17 & 62.30 & 87.25 & 19.63 & 60.66 & 60.97 & 76.29 \\
\bottomrule 
\end{tabular} 
\label{tab:seal_comparison} 
\end{table}

\section{External Tools ViCrop}
\label{app:external}

In this section, we present three automatic question-guided localization methods based on popular off-the-shelf vision-based models, namely CLIP~\cite{clip}, YOLO~\cite{yolo}, and SAM~\cite{kirillov2023segment}. These three methods utilize external vision-based knowledge for the localization process through multimodal encoding, object detection, and semantic segmentation, respectively. See~\cref{tab:externals_time} for their results compared to internal ViCrop methods.

\textbf{CLIP ViCrop.} The intuition of this method is to progressively refine the image towards the region of highest relevance to a given question using CLIP~\cite{clip}. CLIP consists of an image encoder and a text encoder, which are trained on a large dataset of image-caption pairs to map each image (caption) close to its caption (image) and far from all other captions (images). The result is an aligned shared space where various images can be directly compared with various texts. To find the region of interest, given an image-question pair, we first crop the image from the four sides (top, bottom, left, and right) at a cropping ratio of 0.9 to produce four overlapping cropped images. We then use CLIP to assess the semantic similarity between these cropped images and the question. The highest-scoring crop is chosen as the input for the next iteration. This process is repeated for 20 iterations, and the cropped image with the highest CLIP similarity to the question is selected for visual cropping.

\textbf{YOLO ViCrop.} Instead of a progressive approach to finding the region of interest, in this method we select candidate regions based on a state-of-the-art object detection method: YOLOv8~\citep{Jocher_YOLO_by_Ultralytics_2023} pretrained on COCO~\cite{lin2014mscoco}. Using YOLO, we filter out regions that contain no salient objects -- \ie, regions for which CLIP could mistakenly assign high similarity. More concretely, for each question-image pair, we first use YOLO to collect bounding boxes for all predicted objects with confidence higher than 0.25 (the recommended default).\footnote{https://docs.ultralytics.com/modes/predict} Then, for each predicted bounding box, we crop its corresponding image and compute its similarity to the question using CLIP. Finally, the bounding box with the highest similarity score is selected as the region of interest for visual cropping.

\textbf{SAM ViCrop.} A limitation of YOLO is that it only provides bounding boxes corresponding to a fixed number of object classes. To overcome this issue, we use the segment anything model (SAM)~\cite{kirillov2023segment}, which has shown state-of-the-art zero-shot segmentation performance. SAM can provide an extensive set of segmentation masks for each image, thus providing a more granular set of salient candidate regions compared to YOLO. More concretely, for each image-question pair, we feed the image into SAM, which provides an extensive set of segmentation masks corresponding to all objects and object parts. Then, we translate these masks into bounding boxes by computing the smallest bounding box that covers each segmentation mask. Finally, the bounding box with the highest CLIP similarity to the question is selected as the region of interest for visual cropping.

Finally, for each method, we crop the smallest covering square (so that the cropped image is not deformed when resized to the input resolution of the MLLM), and provide it to the MLLM in addition to the original image-question pair (as depicted in~\cref{fig:methods}).

\clearpage
\section{Additional Examples on Model's Predictions}
\label{app:examples}

\begin{figure*}[h]
    \centering
    \includegraphics[trim=0 0 0 0, clip, width=0.99\textwidth]{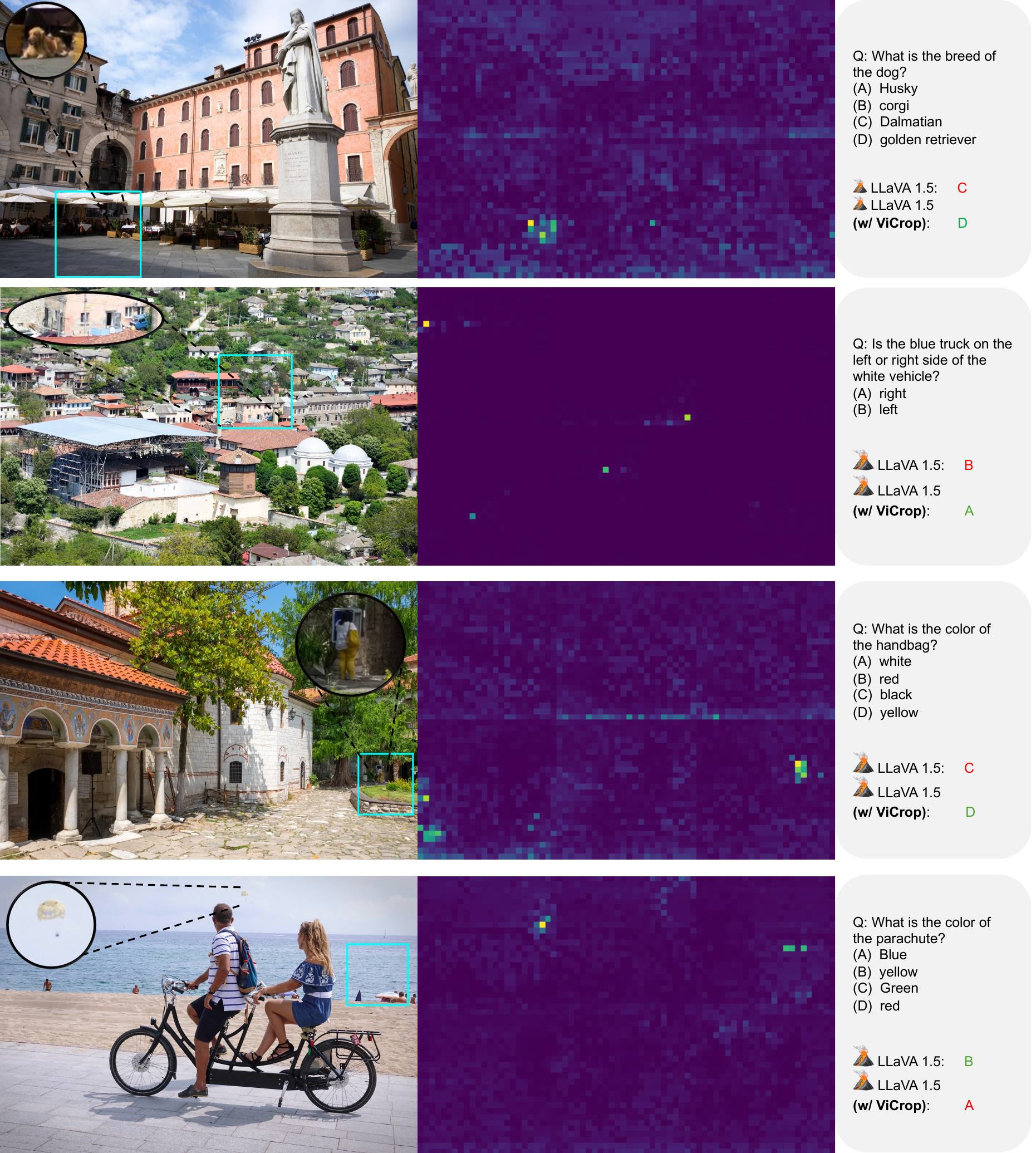}
    \caption{Success (first 3) and failure (last) examples of LLaVA-1.5 (\rel{}) on the V$^*$ benchmark (cyan-colored bounding box shows cropped region by \rel{}; zoom-in insets are displayed for better readability).}
    \label{fig:example_llava_vstar}
\end{figure*}

\begin{figure*}[t]
    \centering
    \includegraphics[trim=0 0 0 0, clip, width=0.99\textwidth]{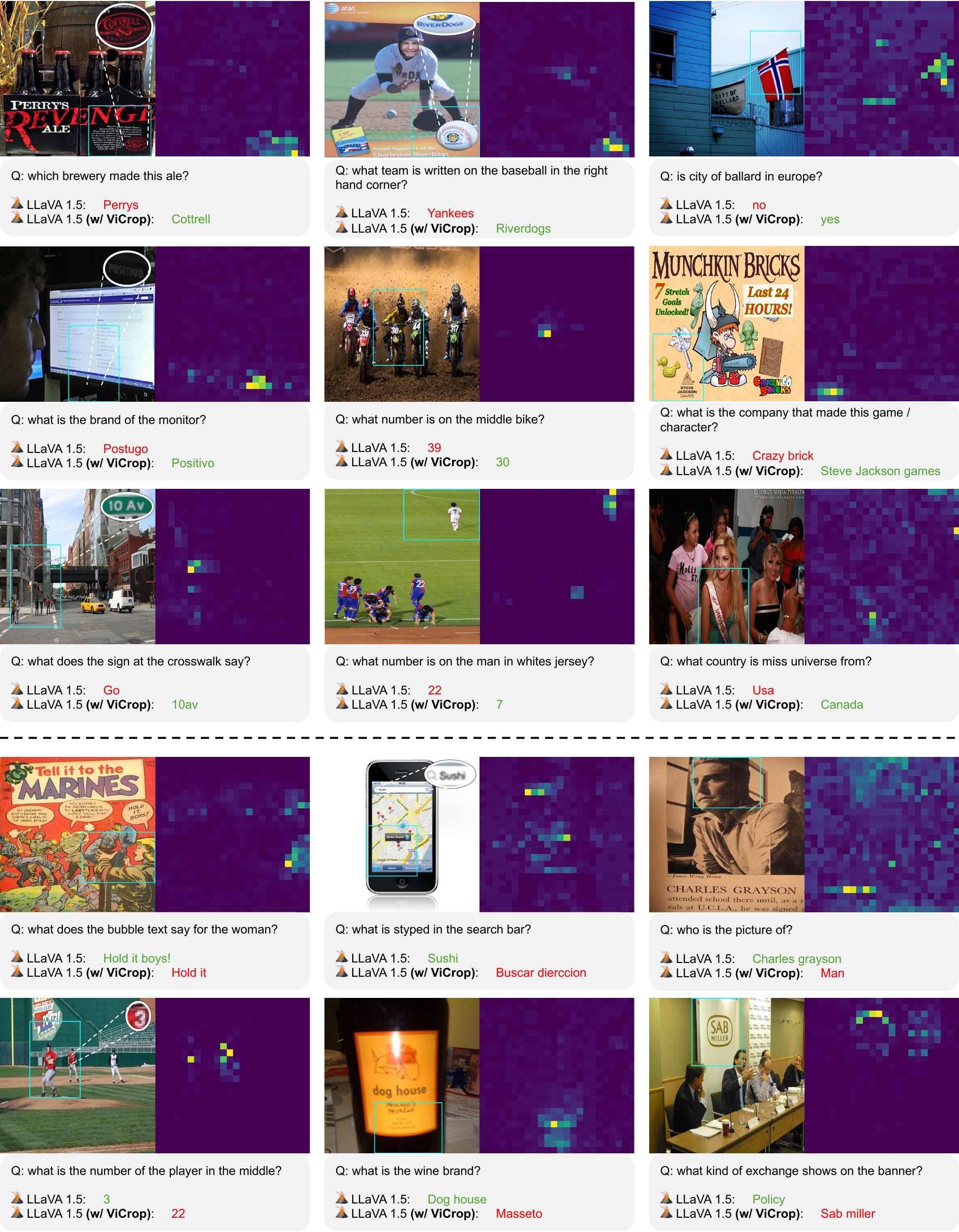}
    \caption{Success (first 9) and failure (last 6) examples of LLaVA-1.5 (\rel{}) on the TextVQA benchmark (cyan-colored bounding box shows cropped region by \rel{}).}
    \label{fig:example_llava_textvqa}
\end{figure*}

\begin{figure*}[t]
    \centering
    \includegraphics[trim=0 0 0 0, clip, width=0.99\textwidth]{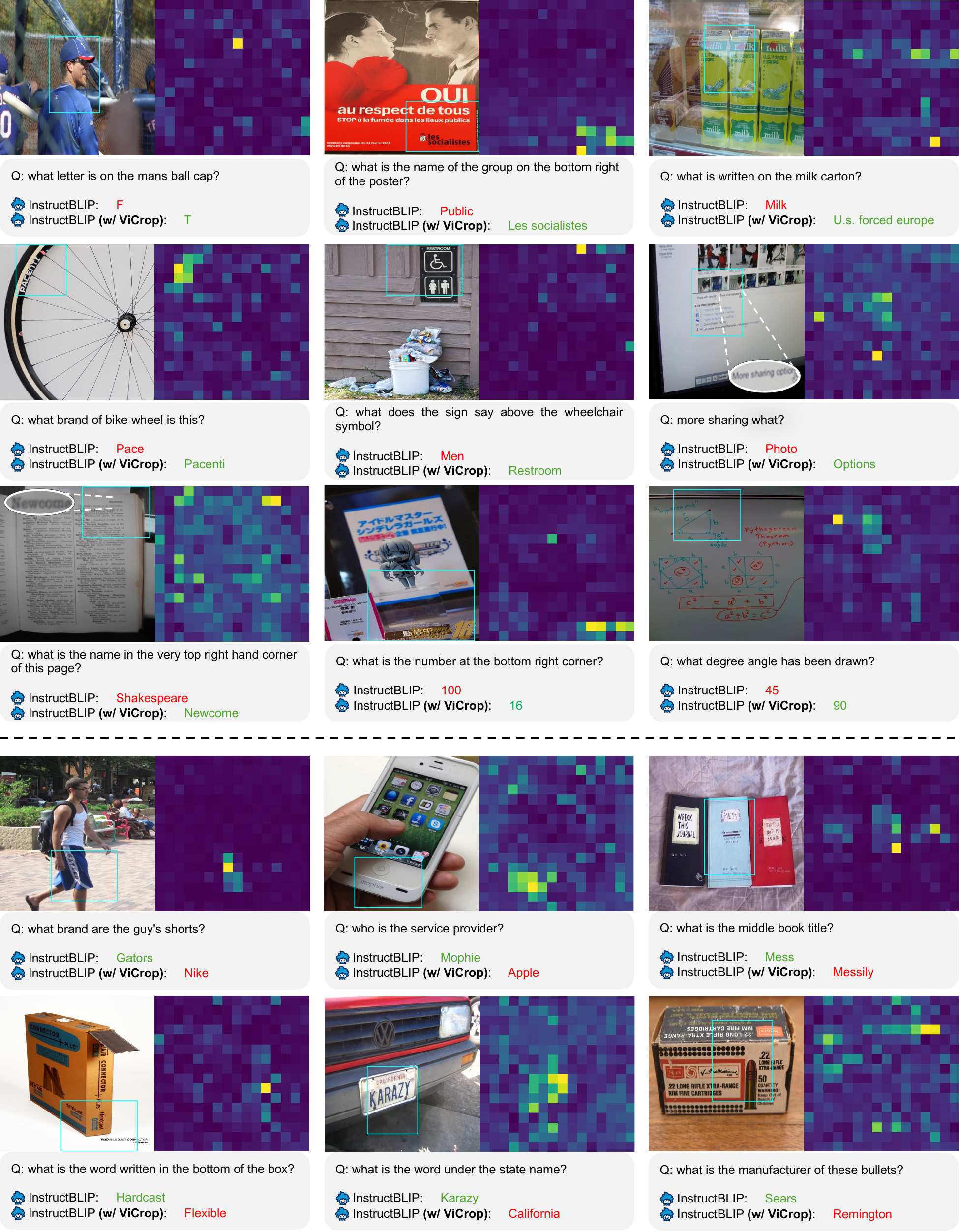}
    \caption{Success (first 9) and failure (last 6) examples of InstructBLIP (\rel{}) on the TextVQA benchmark (cyan-colored bounding box shows cropped region by \rel{}).}
    \label{fig:example_instructblip_textvqa}
\end{figure*}

\end{document}